\documentclass{article}

\usepackage[final,nonatbib]{neurips_2019}

\usepackage[utf8]{inputenc} 
\usepackage[T1]{fontenc}    
\usepackage{hyperref}       
\usepackage{url}            
\usepackage{booktabs}       
\usepackage{amsfonts}       
\usepackage{nicefrac}       
\usepackage{microtype}      

\title{Overcoming Forgetting in Federated Learning on Non-IID Data}

\usepackage[utf8]{inputenc}
\usepackage{graphicx}
\usepackage{amsmath,amssymb}
\usepackage[inline]{enumitem}
\usepackage{xcolor}

\DeclareMathOperator{\diag}{diag}

\DeclareMathOperator{\tr}{tr}

\author{%
  Neta Shoham
    \\
  Edgify \\
  \texttt{neta.shoham@edgify.ai} \\
  \And
  Tomer Avidor \\
  Edgify \\
  \texttt{tomer.avidor@edgify.ai} \\
  \And
  Aviv Keren \\
  Edgify \\
  \texttt{aviv.keren@edgify.ai} \\
  \And
  Nadav Israel \\
  Edgify \\
  \texttt{nadav.israel@edgify.ai} \\
  \And
  Daniel Benditkis \\
  Edgify \\
  \texttt{daniel.benditkis@edgify.ai} \\
  \And
  Liron Mor-Yosef \\
  Edgify \\
  \texttt{liron.moryosef@edgify.ai} \\
  \And
  Itai Zeitak \\
  Edgify \\
  \texttt{itai.zeitak@edgify.ai} \\
}

\begin{document}
\maketitle

\begin{abstract}
We tackle the problem of Federated Learning in the non i.i.d. case, in which local models drift apart, inhibiting learning. Building on an analogy with Lifelong Learning, we adapt a solution for catastrophic forgetting to Federated Learning. We add a penalty term to the loss function, compelling all local models to converge to a shared optimum. We show that this can be done efficiently for communication (adding no further privacy risks), scaling with the number of nodes in the distributed setting. Our experiments show that this method is superior to competing ones for image recognition on the MNIST dataset.
\end{abstract}


\section{Introduction}
Recent years have seen the advent of smart devices and sensors gathering data at the edge and being able to act on that data. The desire to keep data private and other considerations have led the machine learning community to study algorithms for distributed training that do not require sending the data out of the edge devices. Edge devices most often have low networking availability and capacity, which could prohibit training through standard SGD. The {\em Federated Averaging} (FedAvg) algorithm of McMahan et al. \cite{mcmahan2016communication} lets the devices train on their local data for several epochs (using local SGD) before sending the trained model to a central server. The server then aggregates the models and sends the aggregated model back to the devices. This is done iteratively until convergence is achieved.

Federated Learning poses three challenges that make it different from traditional distributed learning. The first one is the number of computing stations, which can be in the hundreds of millions.\footnote {In order to cope with this, it is common practice to select only a subset of devices at every training iteration \cite{mcmahan2016communication}. For simplicity of presentation, we will ignore this method, for which our suggested algorithm can be easily adapted.} The second is much slower communication compared to the inter cluster communication found in data centers. The third difference, on which we focus in this work, is the highly non i.i.d. manner in which the data may be distributed among the devices.

In some real-life cases, Federated Learning has shown robustness to non i.i.d. distribution \cite{ramaswamy2019federated}. There are also recent theoretical results proving the convergence of Federated Learning algorithms \cite{li2019convergence} on non i.i.d. data. It is evident, however, that even in very simple scenarios, Federated Learning on non i.i.d. distributions has trouble achieving good results (in terms of accuracy and the number of communication rounds, as compared to the i.i.d. case) \cite{mcmahan2016communication, wang2019adaptive}.

\subsection{Overcoming Forgetting in Sequential Lifelong Learning and in Federated Learning}
There is a deep parallel between the Federated Learning problem and another fundamental machine learning problem called {\em Lifelong Learning} (and the related {\em Multi-Task Learning}). In Lifelong Learning, the challenge is to learn task 
\(A\), and continue on to learn task \(B\) using the same model, but without "forgetting", without severely hurting the performance on, task \(A\); or in general, learning tasks \(A_1, A_2 \dots \) in sequence without forgetting previously-learnt tasks for which samples are not presented anymore. Besides learning tasks serially rather than in parallel, in Lifelong Learning each task is thus seen only once, whereas in Federated Learning there is no such limitation. But these differences aside, the paradigms share a common main challenge - how to learn a task without disturbing different ones learnt on the same model.

It is not surprising, then, that similar approaches are being applied to solve the Federated Learning and the Lifelong Learning problems. One such example is data distillation, in which representative data samples are shared between tasks \cite{hou2018lifelong,zhao2018federated}. However, Federated Learning is frequently used in order to achieve data privacy, which would be broken by sending a piece of data from one device to another, or from one device to a central point. We therefore seek for some other type of information to be shared between the tasks.

The answer to what kind of information to use may be found in Kirkpatrick et al. \cite{kirkpatrick2017overcoming}. In this work, the authors present a new algorithm for Lifelong Learning - {\em Elastic Weight Consolidation} (EWC). EWC aims to prevent catastrophic forgetting when moving from learning task \(A\) to learning task \(B\). The idea is to identify the coordinates in the network parameters \(\theta\) that are the most informative for task \(A\), and then, while task \(B\) is being learned, penalize the learner for changing these parameters. The basic assumption is that deep neural networks are over-parameterized enough, so that there are good chances of finding an optimal solution \(\theta^*_B\) to task \(B\) in the neighborhood of previously learned \(\theta^*_A\).

In order to control the stiffness of \(\theta\) per coordinate while learning task \(B\), the authors suggest to use the diagonal of the Fisher information matrix  \(\mathcal{I}^*_A=\mathcal{I}_A(\theta_A^*)\) to selectively penalize parts of the parameters vector \(\theta\) that are getting too far from \(\theta^*_{A}\). This is done using the following objective
\begin{equation}\label{eq:deepmind_obj}
\tilde{L}(\theta)=L_B(\theta) + \lambda(\theta-\theta^*_{A})^T  \diag(\mathcal{I}^*_{A})(\theta-\theta^*_{A})
\end{equation}
The formal justification they provide for (\ref{eq:deepmind_obj}) is Bayesian: Let \(D_A\) and \(D_B\) be independent datasets used for tasks \(A\)
and \(B\). We have that
\begin{equation*}
    \log p(\theta|D_A \ \mathrm{and}\  D_B)
    =\log p(D_B|\theta)+\log p(\theta|D_A)-\log p(D_B)  
\end{equation*}\(\log p(D_B|\theta)\) is just the standard likelihood  maximized in the optimization of  \(L_B(\theta)\),
and the posterior \(p(\theta|D_A)\) is approximated with Laplace's method as a Gaussian distribution with expectation \(\theta^*_A\) and covariance \(\mathrm{diag}(\mathcal{I}^*_A)\).

It is also well known that under some regularity conditions, the information matrix approximates the Hessian \(H_L\) of \(L(\theta)\), at \(\theta=\theta^*\) \cite{pronzato2013design}.
By this we get a non Bayesian interpretation of (\ref{eq:deepmind_obj}), 
\begin{equation}\label{eq:hessian_obj}
\tilde{L}(\theta) 
\approx L_B(\theta)+ \frac{1}{2}(\theta-\theta_A^*)^T H_{L_A}(\theta-\theta_A^*) \approx L_B(\theta)+L_A(\theta),
\end{equation}
where \(L(\theta)=L_B(\theta)+L_A(\theta)\) is exactly the loss we want to minimize.  In general, one can learn a sequence of tasks \(A_1 \dots  A_T\). In section \ref{sec:FederatedCurvature} we rely on the above interpretation as a second order approximation in order to construct an algorithm for Federated Learning. We will further show how to implement such an algorithm in a way that preserves the privacy benefits of the standard FedAvg algorithm.

\section{Related Work}
There are only a handful of works that directly try to cope with the challenge of Federated Learning with non i.i.d. distribution. One approach is to just give up the periodical averaging, and reduce the communication by sparsification and quantization of the updates sent to the central point after each local mini batch \cite{sattler2019robust}. In Zhao et al. \cite{zhao2018federated} it was shown that by sharing only a small portion of the data between different nodes, one can achieve a great improvement in model accuracy. However, sharing data is not acceptable in many Federated Learning scenarios.

Somewhat similar to our approach, MOCHA \cite{smith2017federated} links each task with a different parameter \(w_i\in \mathbb{R}^{d \times n}\) and the relation between the tasks is modeled by adding a loss term \(\tr(W\Omega W^T)\), where \(\ W=[w_1,\dots,w_n]\) and \(\Omega\in \mathbb{R}^{n\times n}\). The optimization is done on both \(W\) and \(\Omega\). MOCHA uses a primal-dual formulation in order to solve the optimization problem and thus, unlike our algorithm, is not suitable for deep networks.

Perhaps the closest work to ours is Sahu et al. \cite{Sahu2018FederatedOF}, where the authors present their  FedProx algorithm, which, like our algorithm, also uses  parameter stiffness. However, unlike our algorithm, in FedProx the penalty term is isotropic, \(\frac{1}{2}\mu\Vert \theta - \theta_t \Vert\). DANE  \cite{shamir2014communication} augments FedProx by adding a gradient correction term \(-(\nabla L_i(\theta_{t-1})-\eta\nabla L(\theta_{t-1}) )^T\theta\) to accelerate convergence, but is not robust to non i.i.d. data \cite{Sahu2018FederatedOF,reddi2016aide}. AIDE \cite{reddi2016aide} improves the ability of DANE to deal with non i.i.d. data. However, it does so by using an inexact version of DANE, through a limitation on the amount of local computations.

A recent work \cite{li2019convergence} proves convergence of FedAvg for the non i.i.d. case. It also provides a theoretical explanation for a phenomenon known in practice, of performance degradation when the number of local iterations is too high. This is exactly the problem that we tackle in this work.

\section{Federated Curvature}\label{sec:FederatedCurvature}
In this section we present our adaptation of the EWC algorithm to the Federated Learning scenario. We call it FedCurv (for \textit{Federated Curvature}, motivated by (\ref{eq:hessian_obj})).  We mark by \(S = {\{1 \dots N\}}\) the \(N\) nodes, with the tasks' local datasets \({ \{A_{1},\dots A_N\}} \). We diverge from the FedAvg algorithm and in each round \(t\) we use all the nodes in \(S\) instead of randomly selecting a subset on them. (Our algorithm can easily be extended to select a subset.) At round \(t\) each node \(s \in S \) optimizes the following loss:
\begin{equation}\label{eq:fed_curve}
    \tilde{L}_{t,s}(\theta) = 
    L_{s}(\theta) + 
    \lambda\sum_{{j} \in {S}\setminus {s}}(\theta-\hat{\theta}_{t-1,j})^T\diag(\mathcal{\hat{I}}_{t-1,j})(\theta-\hat{\theta}_{t-1,j}),
\end{equation}
  On each round \(t\), starting from initial point \(\hat{\theta}_t=\frac{1}{N}\sum_{i=1}^N\hat{\theta}_{t-1,i}\), the nodes optimize their local loss by running SGD for \(E\) local epochs. At the end of each round \(t\), each node \(j\) sends to the rest of the nodes the SGD result \( \hat{\theta}_{t,j} \) and \(\diag(\mathcal{\hat{I}}_{t,j})\) (where \(\hat{\mathcal{I}}_{t,j}=\mathcal{I}(\hat{\theta}_{t,j})\)). \( \hat{\theta}_{t,j} \) and \(\diag(\mathcal{\hat{I}}_{t,j})\) will be used for the loss of round \(t+1\). We switched from \(\theta^*\) to \(\hat{\theta}\) to signify that local tasks are optimized for \(E\) epochs and not until they converge (as was the case for EWC). However, (\ref{eq:hessian_obj}) (its generalization to \(N\) tasks) supports using large values of \(E\), so \(\hat{\theta}_{t,j}\approx\theta^*_{t,j}\) and then \(\tilde{L}_{t,j}\approx L\).

\subsection{Keeping Low Bandwidth and Preserving Privacy}
At first glance, 
maintaining all the historical data required by FedCurv might look cumbersome and expensive to store and transmit. It also looks like a sensitive information is passed between nodes. However by careful implementation we can avoid these potential drawbacks. We note that (\ref{eq:fed_curve}) can also be rearranged as
\begin{equation*}
   \tilde{L}_{{t,s}}(\theta) = L_{s}(\theta) + \lambda\theta^T \left[\sum_{{j} \in {S}\setminus {s}}\diag(\hat{\mathcal{I}}_{t-1,j})\right]\theta
   -2\lambda\theta^T\sum_{{j} \in {S}\setminus {s}}\diag(\hat{\mathcal{I}}_{t-1,j})\hat{\theta}_{t-1,j}
    + \mathrm{const} 
\end{equation*}
\paragraph{Bandwidth}The central point needs only to maintain and transmit to the edge node two additional elements, besides \(\theta\), of the same size as \(\theta\),
\begin{equation*}
    u_t=\sum_{{j} \in {S}} \diag(\hat{\mathcal{I}}_{t-1,j})\quad \mathrm{and} \quad
    v_t=\sum_{{j} \in {S}} \diag(\hat{\mathcal{I}}_{t-1,j})\hat{\theta}_{t-1,j}
\end{equation*}
The device can then construct the data needed for the evaluation of \(\tilde{L}\) from \(u_t,\ v_t\) by subtraction. The device \(j\) at time \(t\) needs also two transmit only two additional element at the same size of \(\diag(\hat{\mathcal{I}}_{t-1,j})\) and \(\diag(\hat{\mathcal{I}}_{t-1,j})\hat{\theta}_{t-1,j}\).

\paragraph{Privacy}It should be noted that we only need to send local gradient-related aggregated information (aggregated per local data sample) from the devices to the central point. In terms of privacy, it is not significantly different from the classical FedAvg algorithm. The central point itself, like in FedAvg, needs only to keep globally aggregated information from the devices. We see no reason why secure aggregation methods \cite{bonawitz2016practical} which were successfully applied to FedAvg could not be applied to FedCurv.      

\paragraph{Further potential bandwidth reduction} The diagonal of the Fisher information has been used successfully for parameter pruning in neural networks \cite{lecun1990optimal}. This gives us a straightforward way to save bandwidth by using sparse versions of \(\diag(\hat{\mathcal{I}})\) \(\diag({\mathcal{\hat{I}}})\hat{\theta}\) and even \(\Delta \theta\), as \(\diag(\hat{\mathcal{I}})\) 
provides a natural evaluation for the importance measure of the parameters of \(\hat{\theta}\).  The sparse versions are achieved by keeping only a fraction \(0<q\le 1\) of indices that are related to the \(q\) largest elements of the diagonal of the Fisher information matrix. We have not explored this idea in practice.

\section{Experiments}

We conducted our experiments on a group of 96 simulated devices. We divided the MNIST dataset \cite{lecun1998gradient} into \( 96\times 2\) blocks of homogeneous labels (discarding a small amount of data). We randomly assigned two blocks to each device. We used the CNN architecture from the MNIST PyTorch example \cite{pytorch_mnist}. 

We explored two factors: (1) Learning method - we considered three algorithms, FedAvg, FedProx, and FedCurv (our algorithm); (2) \(E\), the number of epochs in each round, which is of special interest in this work, as our algorithm is designed for large values of \(E\). \(C\), the fraction of devices that participate in each iteration, and \(B\), the local batch size, were kept fixed at \(C=1.0, B=256\). For all the experiments, we have also used a constant learning rate of \(\eta=0.01\). 

FedProx's \(\mu\) and FedCurv's \(\lambda\) values were chosen in the following way: We looked for values that reached 90\% test-accuracy in the smallest number of rounds. We did it by searching on a multiplicative grid using a factor of 10 and then a factor of 2 in order to ensure a minimum. Table \ref{tb:mnist} shows the number of rounds required in order to achieve 95\%, 90\% and 85\% test-accuracy with these chosen parameters. We see that for \(E=50\), FedCurv achieved 90\% test-accuracy three times as fast as the vanila FedAvg algorithm. FedProx also reached 90\% faster than FedAvg. However, while our algorithm achieved 95\% twice as fast as FedAvg, FedProx achieved it two times slower. For \(E=10\), the improvement of both FedCurv and FedProx is less significant, with FedCurv still outperforming FedProx and FedAvg.
 
\begin{figure}[htbp]
\begin{minipage}{.5\textwidth}
  \centering
  \includegraphics[width=1.0\textwidth]{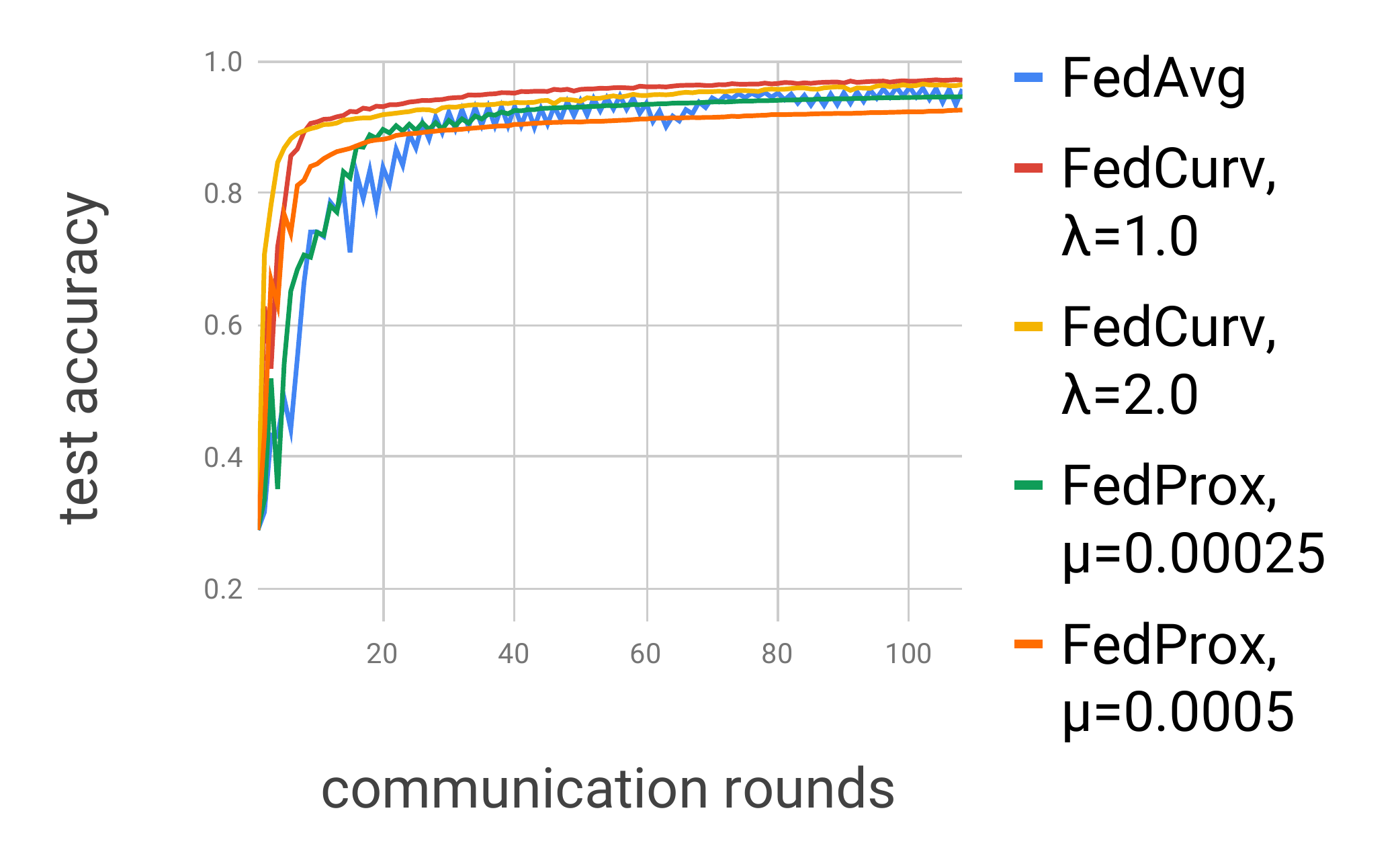}
  \caption{Learning curves, \(E\)=50}
  \label{fig:E50}
\end{minipage}
\begin{minipage}{0.5\textwidth}
\centering
  \includegraphics[width=1.0\textwidth]{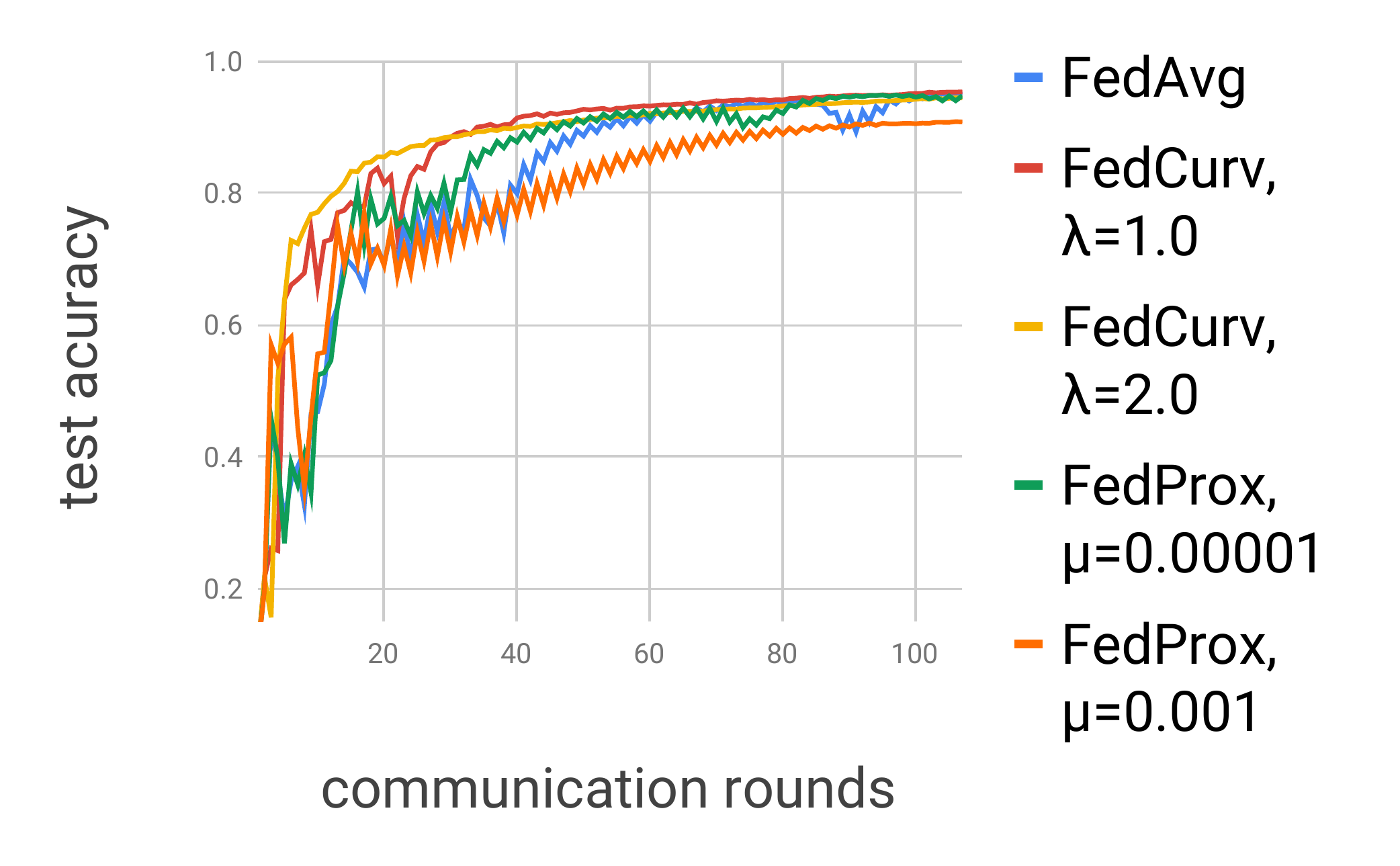}
  \caption{Learning curves, \(E\)=10}
  \label{fig:E10}
\end{minipage}
\end{figure}

 In Figure \ref{fig:E50} and Figure \ref{fig:E10} we can see that both FedProx and FedCurv are doing well at the beginning of the training process. However, while FedCurv provides enough flexibility with \(\theta\) that allows for reaching high accuracy at the end of the process, the stiffness of the parameters in FedProx comes at the expense of accuracy. FedCurv gives more significant improvements for higher values of \(E\) (as does FedProx), as expected by the theory.


\begin{table}
  \caption{Number of rounds to achieve a certain accuracy on Non-IID MNIST}
  \label{tb:mnist}
  \centering
  \begin{tabular}{lllllllll}
    \toprule
    \multicolumn{2}{c}{}&
    \multicolumn{3}{c}{\(E=50\)}        &     
    \multicolumn{1}{c}{}&
    \multicolumn{3}{c}{\(E=10\)}             \\
    \cmidrule(r){3-5}
    \cmidrule(r){7-9}
    Algorithm                &   &0.95   & 0.90 & 0.85&&0.95&0.90&0.85  \\
    \midrule
    FedCurv, \(\lambda=1.0\)    &&  \textcolor{red}{38}  & \textcolor{red}{9}& \textcolor{red}{6}
                                                    &&\textcolor{red}{99}     &\textcolor{red}{35}
                                                               &\textcolor{red}{27} \\
    FedProx, \(\mu=0.00025\)    &&  140  & 22    &16 \\
    FedProx, \(\mu=0.00001\)    &&      & &     &&115   &46&33 \\
    FedAvg                      &&  76   & 30    &22&&106   &51&43\\
    \bottomrule
  \end{tabular}
\end{table}

\section{Conclusion}
This work has provided a novel approach to the problem of Federated Learning on non i.i.d. data. It built on a solution from Lifelong Learning, which uses the diagonal of the Fisher information matrix in order to protect the parameters that are important to each task. The adaptation required modifying that sequential solution (from Lifelong Learning) into a parallel form (of Federated Learning), which a priori involves excessive sharing of data. We showed that this can be done efficiently, without substantially increasing bandwidth usage and compromising privacy. As our experiments have demonstrated, our FedCurv algorithm guards the parameters important to each task, improving convergence.


\clearpage
\bibliographystyle{unsrt}  
\bibliography{references,manual_ref}  
\end{document}